Large Language Models (LLMs) as Traffic Control Systems at Urban Intersections: A New Paradigm

Sari Masri
Arab American University
13 Zababdeh, P.O Box 240 Jenin, Palestine
s.masri3@student.aaup.edu

Huthaifa I. Ashqar*
Civil Engineering Department
Arab American University
13 Zababdeh, P.O Box 240 Jenin, Palestine
Artificial Intelligence Program
Fu Foundation School of Engineering and Applied Science
Columbia University
500 W 120th St, New York, NY 10027, United States
huthaifa.ashqar@aaup.edu

Mohammed Elhenawy
CARRS-Q
Centre for Data Science
Queensland University of Technology
130 Victoria Park Rd, Kelvin Grove QLD 4059, Australia
mohammed.elhenawy@qut.edu.au

*Corresponding Author

# Large Language Models (LLMs) as Traffic Control Systems at Urban Intersections: A New Paradigm

## Abstract

This study introduces a novel approach for traffic control systems by using Large Language Models (LLMs) as traffic controllers. The study utilizes their logical reasoning, scene understanding, and decision-making capabilities to optimize throughput and provide feedback based on traffic conditions in real-time. LLMs centralize traditionally disconnected traffic control processes and can integrate traffic data from diverse sources to provide context-aware decisions. LLMs can also deliver tailored outputs using various means such as wireless signals and visuals to drivers, infrastructures, and autonomous vehicles. To evaluate LLMs ability as traffic controllers, this study proposed a four-stage methodology. The methodology includes data creation and environment initialization, prompt engineering, conflict identification, and fine-tuning. We simulated multi-lane four-leg intersection scenarios and generates detailed datasets to enable conflict detection using LLMs and Python simulation as a ground truth. We used chain-of-thought prompts to lead LLMs in understanding the context, detecting conflicts, resolving them using traffic rules, and delivering context-sensitive traffic management solutions. We evaluated the prformance GPT-mini, Gemini, and Llama as traffic controllers. Results showed that the fine-tuned GPT-mini achieved 83% accuracy and an F1-score of 0.84. GPT-mini model exhibited a promising performance in generating actionable traffic management insights, with high ROUGE-L scores across conflict identification of 0.95, decision-making of 0.91, priority assignment of 0.94, and waiting time optimization of 0.92. This methodology confirmed LLMs' benefits as a traffic controller in real-world applications. We demonstrated that LLMs can offer precise recommendations to drivers in real-time including yielding, slowing, or stopping based on vehicle dynamics. This study demonstrates LLMs transformative potential for traffic control, enhancing efficiency and safety at intersections.

## Introduction

Traffic control systems are essential for maintaining safety and efficiency at urban transportation networks, where a mixed traffic of vehicles, pedestrians, and public transportation vehicles mix at intersections and other conflict points. By managing the flow of traffic through the intersection, these systems aim to reduce congestion, prevent crashes, minimize delays, and save energy to ensure a smoother, safer, and environment-friendly experience for all road users [1], [2], [3]. Effective and timely traffic control is especially crucial in urban areas, where densely populated environments and high mobility demand can quickly lead to congestion, pollution, and safety critical events if not managed well [4], [5]. Traffic signals, signage, and intelligent transportation systems (ITS) are the backbone of this control system, which guide traffic flow based on established algorithms, sensor inputs, and historical data to optimize intersection usage, and increase its efficiency [6], [7].

However, modern urban traffic control systems face significant challenges due to the increasing complexity and dynamism of city traffic patterns [8], [9]. Traditional systems often struggle to adapt to real-time changes in traffic conditions, such as sudden surges in vehicle numbers, diverse modes of transportation, and unpredictable pedestrian and driver behaviors, particularly in mixed traffic scenarios with pedestrians, cyclists, and vehicles [10], [11], [12], [13]. Additionally, optimizing multiple objectives simultaneously such as safety, operational efficiency, and environmental impact further poses more challenges for many conventional traffic control systems [14], which usually lack the ability to balance these goals [15], [16]. Thus, there is a persistent need for more advanced, flexible, and adaptive traffic control solutions that can respond dynamically to real-time data and manage traffic flows through a holistic method. Large Language models (LLMs) have the potential to transform traffic control systems as they have been exposed to massive data, allowing them to enhance their performance of logical reasoning, scene understanding, and decision-making processes [17], [18], [19], [20], [21], [22].

In the context of traffic management, LLMs can handle data efficiently and create a context-appropriate responses and predictions. Previous studies indicated the effectiveness of LLMs in traffic signals management including LLMLight and TrafficGPT [23], [24]Nevertheless, investigating the use of LLMs as traffic controller is still missing in the literature [4]. The contributions of the paper are as follows:

1. Proposing a novel paradigm where LLMs act as dynamic traffic controllers, leveraging their logical reasoning, scene understanding, and decision-making capabilities.
2. Presenting an LLM-based 4D framework for traffic control systems that centralizes traffic control processes traditionally managed by disconnected components.
3. Outlining a four-stage process for integrating LLMs into traffic control, including data creation and initialization, prompt generation using the chain-of-thought method, conflict identification and resolution, and fine-tuning and performance analysis.
4. Evaluating the performance of LLM Models using metrics such as accuracy, precision, recall, F1-score, and ROUGE-L. It identifies the GPT-mini model as the most effective in conflict detection and traffic management tasks.

## LLMs as Traffic Control Systems: A New Paradigm

This study introduces a new paradigm for traffic control system using LLMs as traffic controllers that can serve as practical and efficient alternatives to the current signalized intersection systems. LLMs has shown potential in several applications for their advanced logical reasoning and decision-making capabilities [21], [26], [27], [28], [29]. This study argues that LLMs offer a transformative approach to traffic control, where their inherent flexibility and adaptability can dynamically respond to changing traffic conditions, unlike traditional fixed or adaptive control systems [9], [30], [31]. By deploying LLMs in traffic management, intersections could benefit from more adaptive, context-aware decision-making that addresses real-time traffic flows and diversely mixed traffic types (vehicles, pedestrians, cyclists, and public transportation) and providing tailored-module feedback to drivers with minimal delay [32]. Replacing (or enhancing) existing systems with LLM-based controllers could lead to several advantages including optimized traffic throughput, enhanced response to fluctuating traffic demands, and a reduction in congestion and waiting times [33]. This LLM-driven control could significantly reduce environmental impact by optimizing fuel consumption and minimizing idling times, which might contribute to eco-friendly and sustainable urban mobility solutions.

LLMs may also provide a unique capability to handle multi-objective optimization tasks in traffic management. LLMs can concurrently address and prioritize multiple objectives such as safety, operational efficiency, and fuel consumption [34], [35]. This flexibility enables LLMs to balance factors like speed, timing, and flow with environmental considerations. Using pretrained knowledge and real-time data processing, LLMs can adjust signal timing or alternatively provide direct instructions to the drivers to adjust their speed to decrease emissions and fuel consumption while preserving safety and reducing operational delays. Moreover, the adaptive nature of LLMs allows them to evolve and refine their decision-making based on historical and real-time data, which allow them to be capable of learning optimal traffic patterns and adjusting parameters to maximize overall intersection performance [9], [30].

That said, the adoption of LLMs as traffic controller introduces a new paradigm that expands and enhances traditional definitions and features. Traffic control systems are conventionally defined as the management of transportation system users to ensure conflict-free movement [8], [36]. Traditionally, traffic control at intersections involves various devices including signals, signs, and pavement markings, which operates manually or by rule-based algorithms [37]. LLMs, however, allow for a more integrated and centered approach that can combine control components. Their capacity to interpret and process vast amounts and different types of data as input enables them to provide context-specific guidance to human drivers, infrastructures, and autonomous vehicles [38], [39], [40]. This can adapt to diverse intersection designs, including four-leg intersections, roundabouts, interchanges, and emerging urban intersection models. Moreover, LLMs can communicate instructions through voice, signals, code, visuals, or wirelessly via vehicle-to-infrastructure (V2I) and vehicle-to-vehicle (V2V) technologies [18], [41], [42], which can enhance existing control frameworks.

This study presents LLM as a traffic controller through the concept of *4D traffic control system* (i.e., Detect, Decide, Disseminate, Deploy) as shown in Figure 1 [8], [37], [43]. LLMs could centralize and integrate the disconnected process of detecting and responding to various traffic conditions, which characterizes the traditional traffic controller systems as shown in Figure 1 (a). We argue that LLMs can leverage data from a multitude of detection technologies at the same time such as loop detectors, GPS, video imaging, and DSRC, processing both individual and group trajectories to gain a comprehensive understanding of traffic flows in urban intersections [34]. This detection informs their decision-making process, which allows LLMs to logically reason "where," "when," "who," and "how" to control traffic based on real-time and predictive analysis. Moreover, the integration of multiple objectives in the "Decide" phase, from safety to operational efficiency and energy savings, ensures that LLMs align with specific key goals in traffic management.

In the "Disseminate" and "Deploy" phases, LLMs has the potential to show more versatility. LLMs can interact with both human, machine, and other road infrastructures through multiple communication channels and through different means including voice, visual signals, wireless signals, code, or DSRC. LLMs can also provide timely and effective control outputs that can be binary (stop/go) or non-binary (adjustments to speed or path through logical and actionable instructions to drivers). Moreover, the control outputs generated by LLMs can be modulated based on user type—whether human drivers, pedestrians, or autonomous vehicles—enabling tailored responses that increase compliance and reduce risks associated with complex traffic conditions [26], [28], [38].

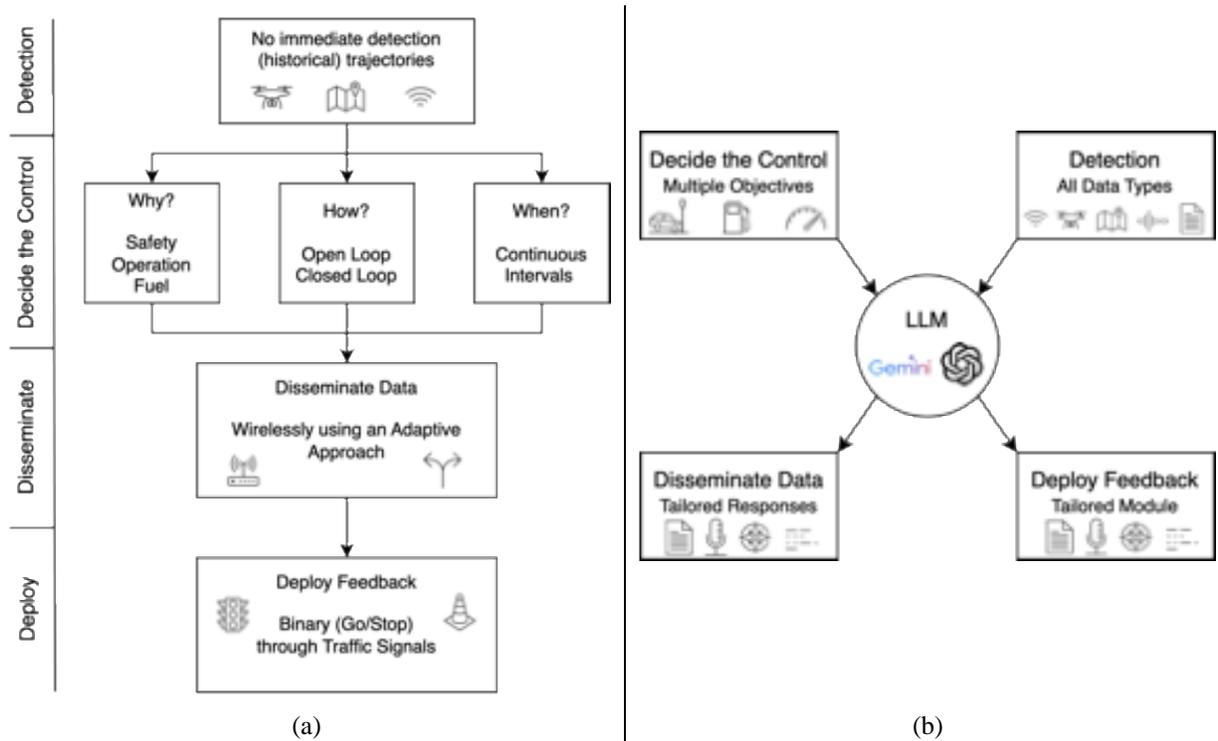

(a)        (b)

*Figure 1: A comparison between (a) the traditional traffic control systems, and (b) the proposed new paradigm of using LLM as a traffic controller. The traditional system uses historical data and structured intervals for decision-making and control dissemination through standard traffic signals. In contrast, the LLM-based system leverages diverse real-time data sources and advanced decision-making capabilities, allowing for adaptive, multi-objective control dissemination and tailored feedback deployment.*

# Background

LLMs has the potential to enable more flexible, adaptable, explainable traffic control systems as well as able to provide actionable feedback to the drivers, traffic engineers, and policymakers.

Previous studies presented the integration of road traffic management solutions using LLMs and Internet of Things (IoT) technologies [31], [33], [44]. Different solutions were introduced such as routing mechanisms, intelligent transportation light solutions, or network traffic management strategies, thereby classifying them [31], [33], [44]. A survey on traffic management with machine and deep learning has been conducted to exemplify the advantages and disadvantages of such techniques [45]. Other surveys highlighted future research directions and give insight into how machine learning and deep learning can help solve traffic management problems [46].

Leveraging LLMs in recent work for traffic prediction involved incorporating sequence and graph embedding layers to obtain features compatible with the input formats of LLMs, followed by utilizing efficient fine-tuning techniques [23].Experiments showed that these frameworks achieved impressive historical sample size and few-shot prediction performance [23]. Innovative combinations of multiple LLMs and traffic foundation models can be an attractive approach for using LLMs with the proper ability to perceive, analyze, and manipulate transportation data [24]. Moreover, other studies proposed a multi-task decision-making system for autonomous driving [47] that utilizes reinforcement learning and LLM sequence modeling to perform effectively in complicated settings [48] such as uncontrolled intersections [49]. Furthermore, it has been shown that using LLMs with input from several sensors, among them cameras and LiDAR, enhances theme perception of transport scenes in a sensible concession of an all-round transportation survey [25]. In the context of advanced connected and automated driving, including LLMs also helps harness traffic management effectiveness [50].

Moreover, the capacity of LLMs to develop a safety case with minimum human effort in traffic management shows and ensures compliance and safety [51]. Upgrading capabilities of prognosis of transportation functions based on LLMs presents a new transformative approach to advance transportation systems [52].

## Methodology

The proposed framework for introducing LLMs as traffic control system and evaluating its performance in real-time is shown in Figure 2. It broadly comprises of four key stages including data creation and initialization, prompt generation, conflict identification, and fine-tuning and model analysis.

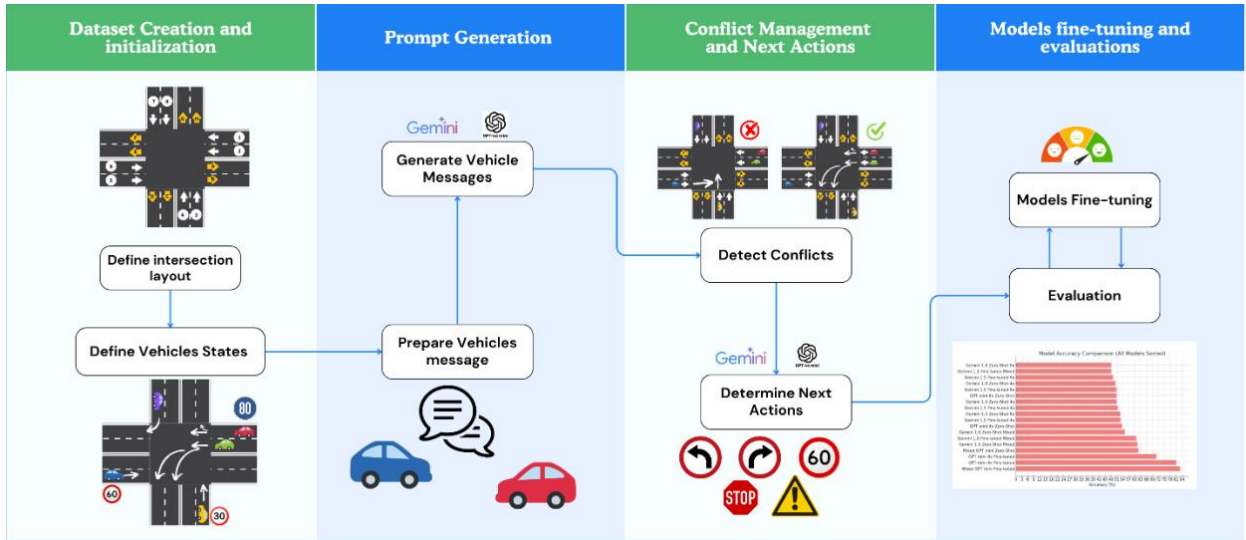

*Figure 2: Flowchart of the proposed framework.*

## Dataset Creation and Initialization

The framework starts with dataset creation and initializations of the environment, where it explains the intersection layout to the LLM including lanes, entering and exiting points, and initial parameters for each vehicle such as speed, initial lane, direction, and destination through the intersection. A traffic dataset was developed to cover intersection scenarios in real-world urban settings. This was achieved by creating a system based on rational thought, calculation methods, and traffic laws. The following section describes the steps involved in producing and generating the actual datasets for the scenarios. The process can be found in Figure 3, which outlines the workflow for scenario preparation and data collection.

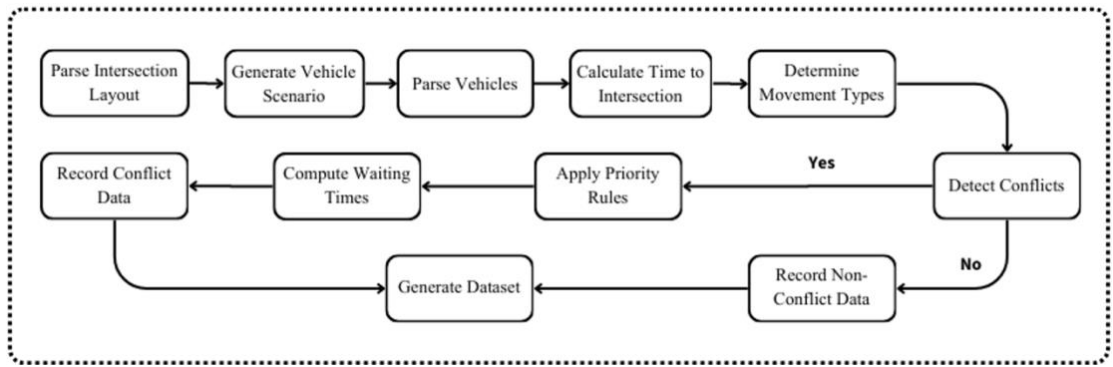

*Figure 3: Workflow for dataset creation system.*

First, the intersection layout is parsed to get information concerning individual lanes and possible destinations from any given lane. Vehicle movement analysis is made more accessible and facilitated by storing the layout in a structured JSON format. Parsing helps in understanding how the intersection is arranged, which is essential when it comes to determining where vehicles will pass and where there may be danger of accidents occurring. We assumed the initial testing to be done on a multi-lane four-legged urban intersection, where vehicles can travel in different directions as shown in Figure 4. Each of its legs has lanes marked for specific purposes, including moving forward only and taking left or rightward turns. This allows traffic from all directions across the 'cross' section with left and right turns facilitated by specific lanes to avoid any conflicts with traffic.

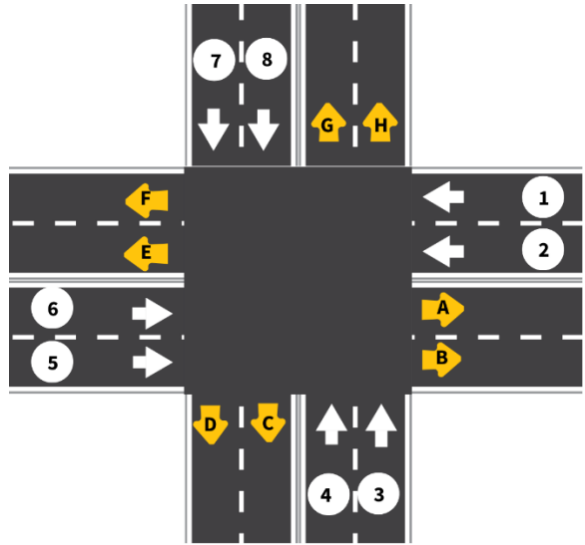

*Figure 4: Multi-lane four-leg intersection as an initial example.*

Table 1 shows the different attributes that was used to define the environment to the LLM. Each instance of the vehicle class represents each vehicle approaching the intersection. The methods in the Vehicle class are used for validating input, calculating time, and determining movement. Input validation is responsible for checking the correctness of all vehicle data attributes, including non-negative speed, distance greater than zero, and valid directions. The time calculation method is meant to determine how much longer it will take this vehicle to reach the intersection, given its current rate of travel as well as how far away it is from there. Consequently, movement determination stands here to classify the vehicle movement into certain types in advance, such as at intersections, lanes, and destinations.

*Table 1: Vehicle attributes defined for the LLM.*

| Attribute | Description |
| --- | --- |
| Vehicle_ID | Unique identifier for the vehicle |
| Lane | The lane number the vehicle is in |
| Speed | Vehicle's speed in km/h |
| Distance_to_Intersection | Distance from the vehicle to the intersection (in meters) |
| Direction | Direction of approach (north, east, south, west) |
| Destination (Egress) | Intended exit lane number from the intersection |

A vehicle is parsed by reading the JSON scenario data of the vehicle and forming an object for each vehicle. To avoid duplicating Vehicle IDs, the system confirms them while ensuring that all requested information is correct. This turns raw scenario data into organized objects, which can be subject to further analysis.

## Prompt Design

In order to properly navigate the frame and utilize LLM features, we created prompts utilizing chain-of-thoughts method. These make it possible for the LLM to undertake a situational analysis, identify probable conflicts and give advice. The prompts were structured to have details on a traffic situation, vehicle conditions and potential conflicts, hence enabling the LLM to make decisions concerning the case. Table 2 illustrates a sample prompt that was applied. This prompt allows the LLM to have background information about traffic and give specific location-based guidance on managing the intersection best.

*Table 2: An example of a used prompt.*

| |
|---|
| You are an Urban Intersection Traffic Conflict Detector, responsible for monitoring a four-way intersection with traffic coming from the north, east, south, and west. Each direction has two lanes guiding vehicles to different destinations: <br> - North: Lane 1 directs vehicles to F and H, Lane 2 directs vehicles to E, D, and C. <br> - East: Lane 3 leads to H and B, Lane 4 leads to G, E, and F. <br> - South: Lane 5 directs vehicles to B and D, Lane 6 directs vehicles to A, G, and H. <br> - West: Lane 7 directs vehicles to D and F, Lane 8 directs vehicles to B, C, and A. <br> Analyze the traffic data from all directions and lanes, and determine if there is a potential conflict between vehicles at the intersection. Respond only with 'Yes' or 'No' for conflict detection. <br> Output: <br> - If there **is a conflict**, provide a report with the following structure: <br> **Conflict Status**: State whether a conflict is detected (e.g., "Conflict detected."). <br> **Conflicts Overview**: Mention the number of conflicts and any vehicles involved (e.g., "Number of conflicts: 1. Involved vehicles: Vehicle V1234 and Vehicle V5678."). <br> **Actions & Decisions**: Summarize any key decisions or actions taken (e.g., "Decisions: Vehicle V5678 must yield to Vehicle V1234."). <br> **Priority Assignment**: List the vehicles and their assigned priorities (e.g., "Vehicle V1234: Priority 1, Vehicle V5678: Priority 2."). <br> **Vehicle Waiting Times**: Provide waiting times for each vehicle (e.g., "Vehicle V1234: 5 seconds, Vehicle V5678: 10 seconds."). <br> **The output format must exactly follow this structure in case of conflict:** |

## Conflict Detection

The creation of vehicle scenarios requires careful consideration of conflict detection, which involves several successive steps. First, the system analyzes whether a particular type of movement may cause intersecting paths between vehicles. For example, vehicles moving in opposite directions without turning may not cross each other's paths at all. Then, it assesses whether any vehicles are likely to arrive at the intersection within a close time interval, potentially creating conflicts. To handle these identified conflicts, the system applies traffic priority rules to determine if a vehicle should yield. Factors such as arrival times and relative positions are taken into account when developing these rules based on movement types—left, straight, or right. The system also calculates waiting times at the junction for lower-priority vehicles, as illustrated in Figure 4, where conflicts and priority decisions, including waiting durations before movement, are captured.

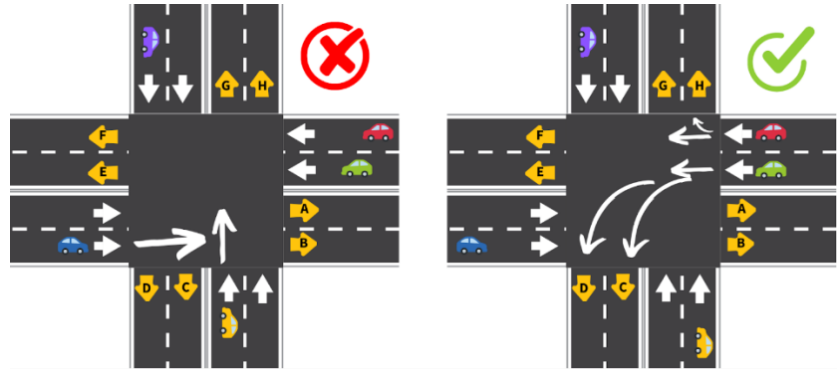

*Figure 5: Comparison of conflict and non-conflict of vehicles.*

The process begins with system initialization, where intersection layout data—essential for understanding possible movements—is read and organized. Then, a random vehicle scenario with various types of vehicles is generated, and a Vehicle object is created for each vehicle. Predefined traffic rules identify conflicts and specify which vehicles should yield, but the LLM goes beyond simple rule application. After evaluating initial conditions, the LLM reviews each detected conflict in real time, providing contextual explanations. It considers factors such as vehicle positions, speeds, and predicted arrival times to recommend specific actions. By offering context-aware driver guidance on yielding decisions and speed adjustments, the LLM enhances traditional rule-based systems, making the interaction more adaptable and human-centered. In cases with no conflicts, scenario data is retained for completeness, while a textual description is generated for human understanding. The procedure continues until all steps are complete, with waiting durations determined for yielding vehicles and conflicts, decisions, and wait times documented.

The system also utilizes standard road traffic priority rules to establish the right of way between conflicting vehicles. Vehicles intending to go straight generally have priority over those planning to turn. For instance, if a vehicle wants to make a left turn, it must wait until all vehicles turning right have cleared the intersection. If two vehicles arrive at the same time with no clear priority, the right-hand rule is applied, where the vehicle on the right has precedence. These rules are essential for resolving conflicts in alignment with common traffic regulations.

As an example, consider a scenario where four vehicles are approaching an intersection. Vehicle V217 is coming from the north in lane 2 at 40 km/h, 80 meters from the intersection, heading toward destination E. Vehicle V218 is approaching from the east in lane 4 at 40 km/h, also 80 meters away, heading toward destination G. Vehicle V219 is approaching from the south in lane 6 at 40 km/h, 80 meters away, heading toward destination A. Lastly, Vehicle V220 is approaching from the west in lane 8 at 40 km/h, 80 meters away, heading toward destination C. In this scenario, the system creates Vehicle objects for V217, V218, V219, and V220, computing each vehicle's time to reach the intersection. Based on the lanes and destinations, it determines that all vehicles are likely making left turns, leading to potential conflicts as they approach simultaneously. Using traffic rules, the system decides which vehicles must yield. For instance, Vehicle V217 may yield to Vehicle V218 if the latter is coming from the right. Waiting times are calculated for each yielding vehicle, and conflict data, including decisions and waiting durations, are recorded and described for human interpretation.

However, in order to make scenarios more interpretable, a system is used to transform them into human-readable textual descriptions. It takes scenario data and pulls out key details concerning every single vehicle. For each vehicle, explanations specify the lane, destination point, speed, distance from the intersection or crossed path, and approach direction. Such descriptions are then aggregated into a coherent whole which outlines what went down in an entire scene. This is illustrated in Table 3.

*Table 3: An example of parsing vehicles data.*

| JSON Input | Input |
|---|---|
| `{`<br>  `"vehicles_scenario": [`<br>    `{`<br>      `"vehicle_id": "V1151",`<br>      `"lane": "2",`<br>      `"speed": 39.995323464891484,`<br>      `"distance_to_intersection": 388.95660041889687,`<br>      `"direction": "north",`<br>      `"destination": "C"`<br>    `},`<br>    `{`<br>      `"vehicle_id": "V5173",`<br>      `"lane": "8",`<br>      `"speed": 68.0915930855088,`<br>      `"distance_to_intersection": 150.82949998592466,`<br>      `"direction": "west",`<br>      `"destination": "B"`<br>    `},`<br>    `{`<br>      `"vehicle_id": "V8617",`<br>      `"lane": "1",`<br>      `"speed": 43.411746756299856,`<br>      `"distance_to_intersection": 180.7639436593828,`<br>      `"direction": "north",`<br>      `"destination": "F"`<br>    `},`<br>    `{`<br>      `"vehicle_id": "V2618",` | Vehicle V7155 is in lane 2, moving north at a speed of 30.86 km/h, and is 88.54 meters away from the intersection, heading towards D. Vehicle V6439 is in lane 3, moving east at a speed of 53.37 km/h, and is 107.50 meters away from the intersection, heading towards B. Vehicle V5182 is in lane 7, moving west at a speed of 47.69 km/h, and is 94.67 meters away from the intersection, heading towards D. Vehicle V2432 is in lane 1, moving north at a speed of 46.17 km/h, and is 74.59 meters away from the intersection, heading towards H. |
| ```json
    "lane": "4",
    "speed": 63.24744202519462,
    "distance_to_intersection": 366.3390574707282,
    "direction": "east",
    "destination": "F"
   }
  ]
}
``` | |
| **JSON Output**<br>```json
{
 "is_conflict": "yes",
 "number_of_conflicts": 4,
 "places_of_conflicts": [
  "intersection",
  "intersection",
  "intersection",
  "intersection"
 ],
 "conflict_vehicles": [
  {
   "vehicle1_id": "V7155",
   "vehicle2_id": "V6439"
  },
  {
   "vehicle1_id": "V7155",
   "vehicle2_id": "V5182"
  },
  {
   "vehicle1_id": "V6439",
   "vehicle2_id": "V2432"
  },
  {
   "vehicle1_id": "V5182",
   "vehicle2_id": "V2432"
  }
 ],
 "decisions": [
  "Potential conflict: Vehicle V7155 must yield to Vehicle V6439",
  "Potential conflict: Vehicle V7155 must yield to Vehicle V5182",
  "Potential conflict: Vehicle V6439 must yield to Vehicle V2432",
  "Potential conflict: Vehicle V5182 must yield to Vehicle V2432"
 ],
 "priority_order": {
  "V2432": 1,
  "V5182": 2,
  "V6439": 3,
  "V7155": 4
 },
``` | **Output**<br>**Conflict Status**: Conflict detected.<br>**Conflicts Overview**: Number of conflicts: 4. Involved vehicles: Vehicle V7155 and Vehicle V6439, Vehicle V7155 and Vehicle V5182, Vehicle V6439 and Vehicle V2432, Vehicle V5182 and Vehicle V2432.<br>**Actions & Decisions**: Decisions: Potential conflict: Vehicle V7155 must yield to Vehicle V6439, Potential conflict: Vehicle V7155 must yield to Vehicle V5182, Potential conflict: Vehicle V6439 must yield to Vehicle V2432, Potential conflict: Vehicle V5182 must yield to Vehicle V2432<br>**Priority Assignment**: Vehicle V2432: Priority 1, Vehicle V5182: Priority 2, Vehicle V6439: Priority 3, Vehicle V7155: Priority 4.<br>**Vehicle Waiting Times**:<br>  - Vehicle V2432: 0 seconds<br>  - Vehicle V5182: 1 seconds<br>  - Vehicle V6439: 3 seconds<br>  - Vehicle V7155: 2 seconds |

```
  "waiting_times": {
    "V2432": 0,
    "V5182": 1,
    "V6439": 3,
    "V7155": 2
  }
}
```

## Model Selection and Fine-Tuning

We tested our novel paradigm (i.e., using LLM as a traffic controller) on three different models including GPT-mini, Gemini, and Llama. They were chosen for their ability to solve natural language problems and their potential to understand traffic situations in real time under changing environments. We tested the models using two learning methods including zero-shot and fine-tuning. This allowed us to analyze their effectiveness in different types of tasks.

The GPT-model mini has been scaled down for it to process faster than the bigger ones and was fine-tuned on three datasets; the first one being a four-vehicle dataset while the second contained eight vehicles' data such as mixed data with examples ranging between two and eight cars in each scenario. This mixed dataset was created based on the assumption that performance would improve over a wide range of traffic complexities.

The Gemini models were also tested. Like GPT-mini, they are transformer-based, but their training methods and design differ. This study used two versions of the Gemini model: Gemini 1.0, evaluated in both fine-tuned and zero-shot conditions, and Gemini 1.5, a more advanced version with improved capabilities for scenario understanding and conflict resolution. Additionally, two versions of the Llama model, Llama-3.1-8B and Llama-3.1-70B.

For fine-tuned models, the training and validation datasets were used to customize the models to the specific task of conflict detection in urban intersections. On the other hand, zero-shot models were not fine-tuned for the specific task and were evaluated directly on the testing dataset. This approach tested the models' ability to generalize to new tasks without specialized training, which offeres insights into their base knowledge in urban traffic control systems.

We examined the performance of these models across both fine-tuned and zero-shot settings. We aimed to determine how effectively LLMs can adapt to real-world traffic management tasks. Thus, each LLM models were analyzed using classification metrics such as accuracy, precision, recall, and F1-score, alongside their ability to generate detailed analyses of traffic scenarios, measured through ROUGE-L scores. We tested the outputs of LLMs using a simulation model that was built on Python.

# Analysis and Results

## Logical Reasoning Results

In this section, we present the results of using LLM to identify conflicts in intersection scenarios. Specifically, evaluation was conducted on different versions of GPT-mini, Gemini, and Llama under both fine-tuned and zero-shot conditions. It must be noted that the datasets were designed to consist of varying numbers of cars to evaluate how well these models generalize within different traffic conditions.

Depending on the number of vehicles involved, datasets were divided. The 4-vehicles dataset involves situations with just four vehicles on the road intersection, while the 8-vehicles dataset is all about having eight cars in similar spots. We also used another mixed-vehicles dataset, in which different numbers of vehicle-based scenarios ranging between 2 and 8 were fed to the LLM. They are organized into three groups of 10,000 scenarios in total including 7,000 that was catered for training purposes, 2,000 available for testing, and 1,000 reserved for validation purposes. Zero-shot evaluation was only performed using test datasets. Test datasets were used for assessment of fine-tune models, and training and validation were used for fine-tuning as well. The performance of the LLM in detecting conflicts at the defined intersection is summarized in Table 4.

*Table 4: Performance of the LLM in detecting conflicts.*

| Model | Learning Method | Scenario | Accuracy | Precision | Recall | F1-Score |
|---|---|---|---|---|---|---|
| GPT-4o-mini | fine-tuning | mixed-vehicle | **83.0** | **0.83** | **0.85** | **0.84** |
| | fine-tuning | 4-vehicle | **81.0** | **0.80** | **0.83** | **0.82** |
| | fine-tuning | 8-vehicle | **71.0** | **0.70** | **0.74** | **0.72** |
| | zero-shot | mixed-vehicle | 61.9 | 0.59 | 0.62 | 0.60 |
| | zero-shot | 4-vehicle | 53.6 | 0.53 | 0.54 | 0.53 |
| | zero-shot | 8-vehicle | 50.8 | 0.50 | 0.51 | 0.42 |
| Gemini 1.5 | zero-shot | 8-vehicle | 52.8 | 0.55 | 0.77 | 0.62 |
| | fine-tuning | 4-vehicle | 51.5 | 0.51 | 0.48 | 0.50 |
| | zero-shot | 4-vehicle | 50.9 | 0.51 | 0.54 | 0.52 |
| | fine-tuning | 8-vehicle | 49.1 | 0.49 | 0.50 | 0.49 |
| | fine-tuning | mixed-vehicle | 48.2 | 0.49 | 0.72 | 0.58 |
| | zero-shot | mixed-vehicle | 61.4 | 0.62 | 0.60 | 0.61 |
| Gemini 1.0 | fine-tuning | mixed-vehicle | 60.9 | 0.61 | 0.62 | 0.61 |
| | zero-shot | mixed-vehicle | 55.1 | 0.57 | 0.42 | 0.49 |
| | fine-tuning | 4-vehicle | 52.9 | 0.55 | 0.73 | 0.61 |
| | fine-tuning | 8-vehicle | 50.8 | 0.60 | 0.05 | 0.09 |
| | zero-shot | 4-vehicle | 50.4 | 0.50 | 0.58 | 0.54 |
| | zero-shot | 8-vehicle | 48.2 | 0.47 | 0.48 | 0.45 |
| Llama-3.1-8B-Instruct | zero-shot | mixed-vehicle | 50.4 | 0.52 | 0.50 | 0.37 |
| | fine-tuning | 4-vehicle | 49.6 | 0.49 | 0.50 | 0.43 |

| Llama-3.1-70B-Instruct | fine-tuning | 4-vehicle | 51.5 | 0.51 | 0.51 | 0.51 |

The GPT-mini fine-tuned model achieved the highest accuracy of 83% on the mixed-vehicle scenario, indicating its superior ability to generalize across scenarios with varying numbers of vehicles. Fine-tuning on a diverse dataset likely contributed to its robust performance. The high recall of 0.85 suggests that the model effectively identifies actual conflict scenarios, which is critical in applications where overlooking a conflict can have severe consequences. The GPT-mini 4-vehicle fine-tuned model also performed well, with an accuracy of 81%. This demonstrates that even when trained on scenarios with a fixed number of vehicles, the model can achieve high performance, although slightly lower than the mixed dataset counterpart. The reduction in accuracy compared to the mixed model may be due to overfitting to scenarios with only four vehicles, limiting its ability to generalize to more varied situations. The GPT-mini 8-vehicle Fine-tuned model's accuracy dropped to 71%, indicating that scenarios with a higher number of vehicles pose a greater challenge.

The complexity introduced by additional vehicles may require more sophisticated modeling or larger datasets to capture the nuances of interactions in such scenarios. Zero-shot models generally underperformed compared to their fine-tuned counterparts. The Mixed GPT-mini Zero-Shot model achieved an accuracy of 61.9%, highlighting the importance of fine-tuning for specialized tasks. Without task-specific training, models struggle to capture the intricacies of conflict detection in intersection scenarios. The Gemini models consistently exhibited lower performance across both fine-tuned and zero-shot settings. The highest accuracy among them was 61.4% for the Gemini 1.5 Zero-Shot Mixed model. This suggests that the Gemini architecture may not be as well-suited for this classification task or that the fine-tuning process was less effective compared to GPT-mini models. The Meta-Llama models, particularly the Meta-Llama-3.1-70B Fine-tuned on 4-vehicle, showed moderate improvements with fine-tuning, achieving a balanced F1-score of 0.51 and an accuracy of 51.5%. This indicates some potential for scalability with larger parameter sizes; however, they still trail behind the GPT-mini models, likely due to a lack of optimization for complex traffic conflict detection. The comparison underscores the effectiveness of fine-tuning on diverse datasets and highlights the suitability of the GPT-mini architecture for nuanced traffic scenarios, as shown in Figure 6.

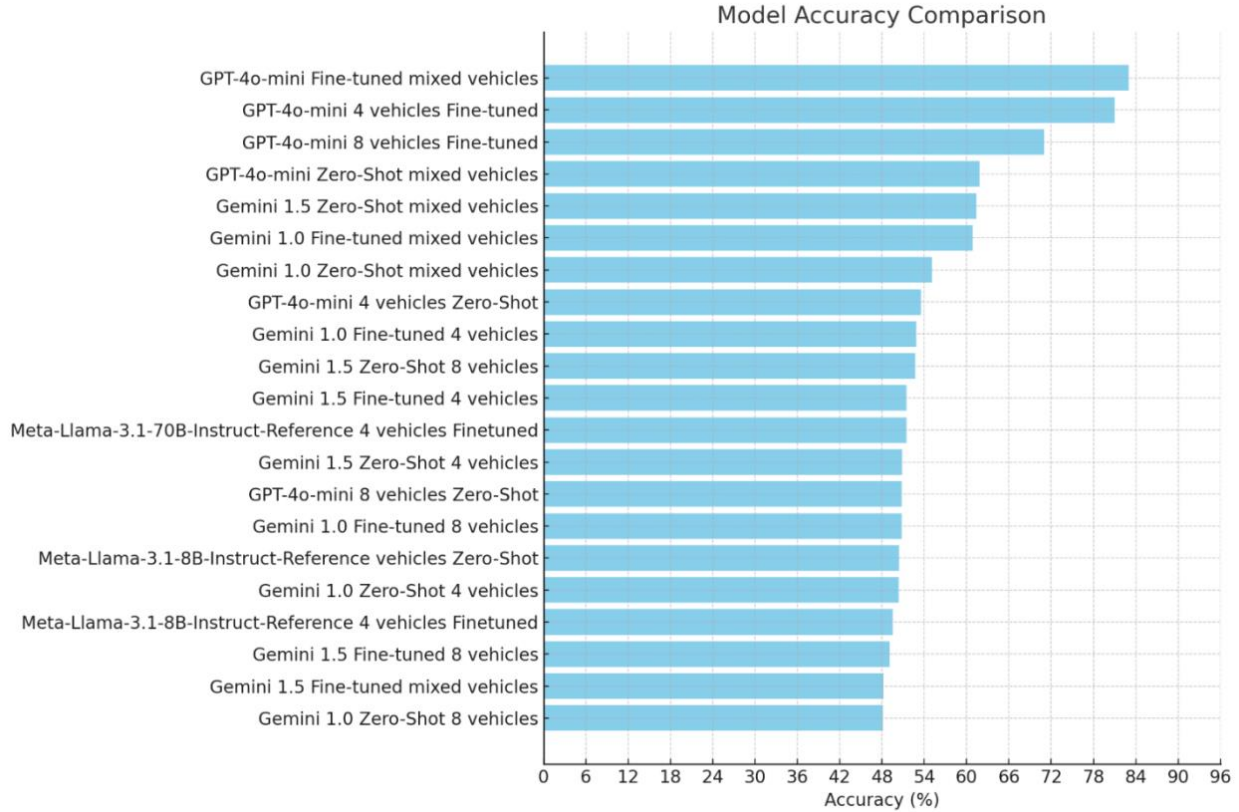

*Figure 6: Accuracy comparison across various LLMs, learning methods, and vehicle scenarios.*

To understand the performance of the fine-tuned GPT-mini models, confusion matrices were analyzed for the best-performing models. Figure 7 presents the confusion matrices for the GPT-mini models in different scenarios. The mixed-vehicle fine-tuned GPT-mini model exhibits the best performance among the three, with a higher number of true positives (820) and true negatives (849), and lower false negatives (180) and false positives (151). This indicates that the model is proficient at correctly identifying both conflict and non-conflict scenarios. Comparing the fine-tuned GPT-mini model for 4-vehicle to the mixed model, we observe slightly lower true positives (797) and true negatives (830), and higher false negatives (203) and false positives (170). This suggests that the model trained exclusively on four-vehicle scenarios may not generalize as effectively as the mixed model, leading to a higher rate of misclassification. The 8-vehicle fine-tuned GPT-mini model shows a decrease in performance compared to the mixed and four-vehicle models, with a higher number of false negatives (324) and false positives (255). This reinforces the earlier observation that scenarios with more vehicles introduce complexity that challenges the model's ability to accurately classify conflict scenarios.

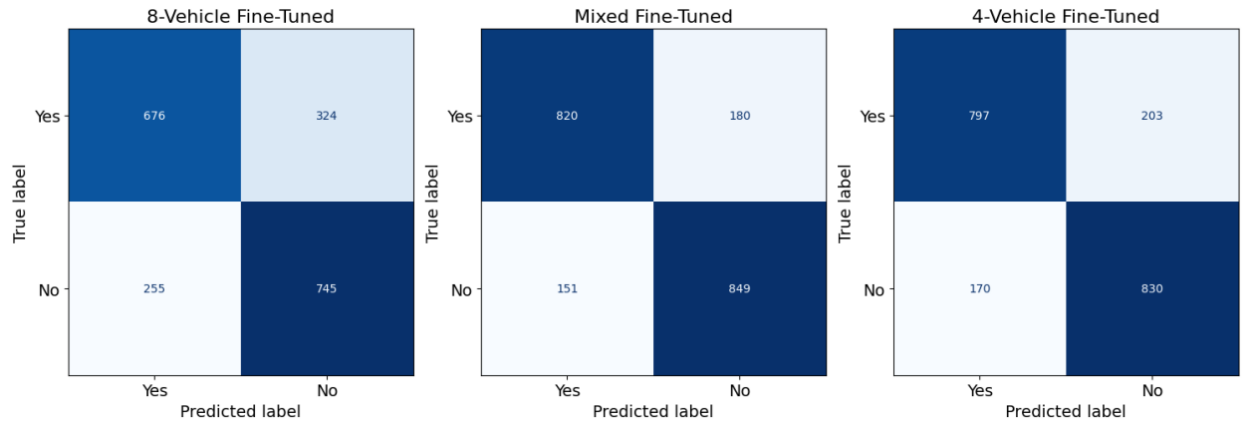

*Figure 7: Confusion matrix for fine-tuned GPT-4 Mini across different vehicle scenarios.*

## Decision-Making and Feedback Results

In this section, we tested the decision-making and feedback that the fine-tuned GPT-mini model provided to the drivers on the 4-vehicle and mixed-vehicle scenarios. Evaluation was conducted using ROUGE-L score based on four measurements including identifying conflicts, decisions, priority assignments, and waiting times. The truth values were found using Python simulation and was validated by traffic engineering experts. Firstly, identifying conflicts involves detecting potential points where vehicles' paths may intersect or where their arrival times may lead to a collision risk at an intersection. Secondly, decisions refer to determining actions for each vehicle in a conflict scenario, such as whether a vehicle should yield, proceed, or adjust its speed to avoid collision. Third, priority assignments involve assigning right-of-way to vehicles based on traffic rules, such as giving priority to vehicles going straight over those making turns or applying the right-hand rule. Finally, waiting times measures the amount of time a vehicle must wait at an intersection before it can proceed, usually due to lower priority or yielding requirements in a conflict situation.

Figure 8 presents the ROUGE-L scores for various. In the 4-vehicle scenarios, the fine-tuned GPT-mini model achieved high ROUGE-L scores across all components, with the conflicts overview scoring 0.95. This indicates that the model can accurately and coherently summarize conflicts in simpler scenarios. The decisions component scored 0.91, demonstrating the model's ability to interpret and explain the decisions made based on traffic rules. The high scores in priority assignment (0.94) and vehicle waiting times (0.92) further showcase the model's proficiency in assigning priorities and calculating waiting times for vehicles. For mixed-vehicle scenarios, the ROUGE scores are slightly lower but still strong. The conflicts overview scored 0.84, and the decisions component scored 0.81. These results suggest that as the complexity of the scenarios increases with varying numbers of vehicles, the model faces more challenges in summarizing conflicts accurately. However, the model maintains respectable performance, indicating its ability to handle complex scenarios effectively. The priority assignment and vehicle waiting times components scored 0.87 and 0.86, respectively, reflecting the model's competence in applying traffic rules and calculating necessary delays even in more complex situations. The consistently high scores in the decisions component across both scenario types underscore the model's strong grasp of traffic rules and its ability to apply them in diverse situations. The slight decline in scores for mixed-vehicle scenarios highlights the increased complexity but also shows that the model adapts reasonably well. These results demonstrate that the fine-tuned GPT-mini model excels not only in binary classification of conflict scenarios but also in generating detailed, coherent analyses. The model's ability to maintain high performance across key components, even in complex scenarios, is promising for practical applications where detailed explanations are necessary.

The high ROUGE scores in critical components indicate that the model can effectively support traffic management tasks by providing accurate conflict detection and actionable decisions. In real-world applications, such capabilities can enhance traffic flow efficiency and safety at intersections by enabling proactive measures to prevent accidents and reduce congestion.

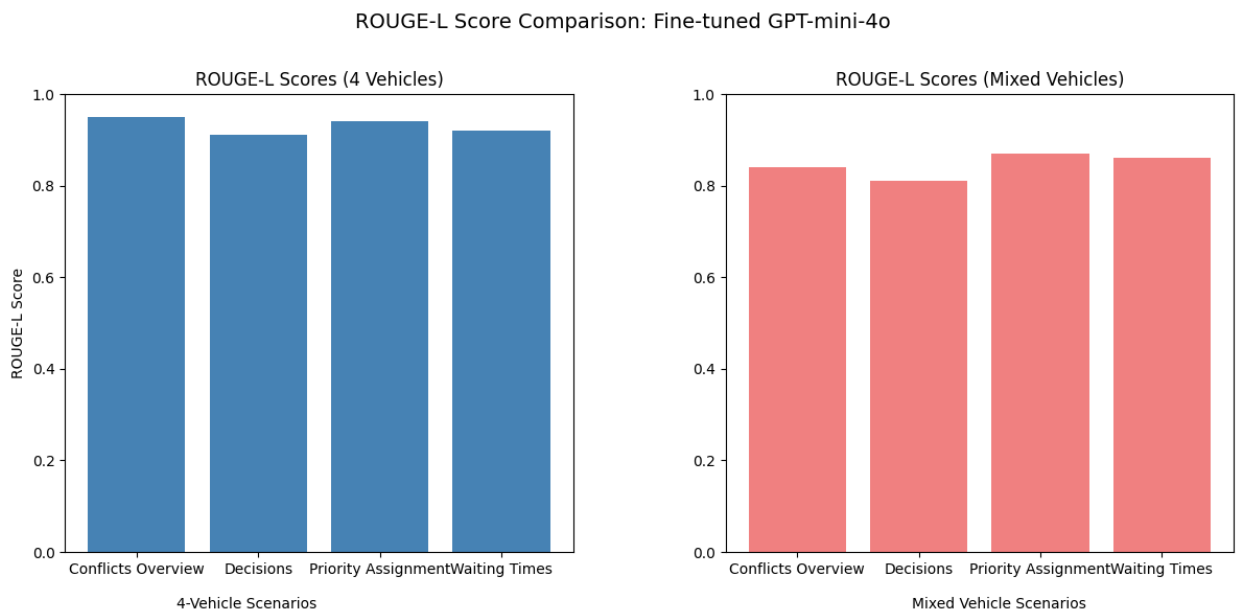

*Figure 8: ROUGE score for the based on identifying conflicts, decisions, priority assignments, and waiting times.*

The fine-tuned GPT model's outputs in Table 5 showcases its capability to identify potential conflicts and provide precise recommendations for vehicles approaching an intersection. In scenarios where conflicts are detected, such as between Vehicles V4625 and V1909 or Vehicles V7019 and V5264, the model suggests actionable measures like yielding or slowing down based on the vehicles' distances, speeds, and intended paths, prioritizing safety and minimizing collision risks. For instance, it recommends Vehicle V1909 yield to Vehicle V4625 due to the latter's advantage in distance and suggests Vehicle V7019 yield to Vehicle V5264 to avoid intersection congestion. In contrast, the model advises all vehicles to continue without adjustment for conflict-free situations, accurately discerning safe scenarios. This approach extends to more complex cases, such as between northbound and westbound vehicles V9867 and V6070, where the model proposes a slight yield from V9867 to ensure safe passage, demonstrating its nuanced understanding of cooperative traffic flow. Combining conflict detection with tailored, real-time guidance, the model displays a structured and responsive framework for intersection management, supporting automated safety and efficiency in multi-vehicle environments.

*Table 5: Examples of outputs of LLM's decision-making and feedback to the drivers.*

| Scenario Input | Fine-tuned GPT-Mini Output |
|---|---|
| Vehicle V7406 is in lane 7, moving west at a speed of 27.59 km/h, and is 182.12 meters away from the intersection, heading towards D. Vehicle V4625 is in lane 3, moving east at a speed of 49.97 km/h, and is 189.12 meters away from the intersection, heading towards B. Vehicle V1909 is in lane 8, moving west at a speed of 75.90 km/h, and is 263.27 meters away from the intersection, heading towards C. Vehicle V2327 is in lane 8, moving west at a speed of 26.36 km/h, and is 445.70 meters away from the intersection, heading towards A. | **Conflict Status**: Yes **Conflict Analysis**: A conflict is detected between Vehicle V4625 and Vehicle V1909. Both vehicles are approaching the same intersection from opposite directions, and they are likely to reach the intersection at the same time due to their respective speeds and distances. **Recommendations**: To prevent a potential collision, it is recommended that Vehicle V1909 yield to Vehicle V4625, as V4625 has a slight advantage in distance. **Next Actions for Each Vehicle**: - Vehicle V4625: Continue straight towards the intersection - Vehicle V1909: Yield and prepare to adjust speed or trajectory if necessary |
| Vehicle V1975 is in lane 2, moving north at a speed of 61.28 km/h, and is 431.49 meters away from the intersection, heading towards D. Vehicle V7019 is in lane 4, moving east at a speed of 50.89 km/h, and is 195.12 meters away from the intersection, heading towards E. Vehicle V5264 is in lane 6, moving south at a speed of 73.17 km/h, and is 207.91 meters away from the intersection, heading towards G. Vehicle V8370 is in lane 3, moving east at a speed of 25.37 km/h, and is 393.16 meters away from the intersection, heading towards H. | **Conflict Status**: Yes **Conflict Analysis**: A potential conflict is detected between vehicles V7019 and V5264. Vehicle V7019 is approaching the intersection, while Vehicle V5264 is already entering the intersection; both are on a collision course. **Recommendations**: It is recommended that Vehicle V7019 yield to Vehicle V5264 to avoid conflict. **Next Actions for Each Vehicle**: - Vehicle V1975: No action - Vehicle V7019: Slow down and yield - Vehicle V5264: Continue at current speed - Vehicle V8370: No action |
| Vehicle V5009 is in lane 6, moving south at a speed of 45.37 km/h, and is 217.81 meters away from the intersection, heading towards H. Vehicle V3191 is in lane 6, moving south at a speed of 68.86 km/h, and is 298.54 meters away from the intersection, heading towards H. Vehicle V6111 is in lane 4, moving east at a speed of 78.52 km/h, and is 106.46 meters away from the intersection, heading towards E. Vehicle V7721 is in lane 7, moving west at a speed | **Conflict Status**: No **Conflict Analysis**: No conflicts detected among the vehicles. **Recommendations**: None **Next Actions for Each Vehicle**: - V5009: Continue moving south - V3191: Continue moving south - V6111: Continue moving east - V7721: Continue moving west |
| of 44.52 km/h, and is 309.51 meters away from the intersection, heading towards F. | |
| Vehicle V9867 is in lane 2, moving north at a speed of 65.73 km/h, and is 270.95 meters away from the intersection, heading towards C. Vehicle V6094 is in lane 6, moving south at a speed of 61.93 km/h, and is 344.02 meters away from the intersection, heading towards G. Vehicle V6070 is in lane 7, moving west at a speed of 59.35 km/h, and is 296.58 meters away from the intersection, heading towards F. Vehicle V2692 is in lane 2, moving north at a speed of 46.17 km/h, and is 118.97 meters away from the intersection, heading towards E. | **Conflict Status**: Yes **Conflict Analysis**: There is a conflict between Vehicle V9867 (northbound) and Vehicle V6070 (westbound). V9867 is approaching the intersection from the north at a speed of 65.73 km/h and is positioned 270.95 meters away, while V6070 is approaching from the west at a speed of 59.35 km/h and has a distance of 296.58 meters to the intersection. The distance to the intersection for both vehicles is similar, which can lead to an intersection conflict. **Recommendations**: The drivers of both vehicles should communicate well and gauge each other's positions. It may be best for Vehicle V9867 to yield slightly to allow for safer navigation through the intersection. **Next Actions for Each Vehicle**: - Vehicle V9867: Yield slightly, if necessary, to allow V6070 to proceed safely through the intersection. - Vehicle V6070: Proceed without delay, being aware of V9867's proximity. - Vehicle V6094: No action needed concerning the conflict. - Vehicle V2692: No action needed concerning the conflict. |
| Vehicle V6758 is in lane 4, moving east at a speed of 61.16 km/h, and is 413.33 meters away from the intersection, heading towards F. Vehicle V5284 is in lane 2, moving north at a speed of 59.44 km/h, and is 248.61 meters away from the intersection, heading towards D. Vehicle V6460 is in lane 5, moving south at a speed of 73.17 km/h, and is 363.19 meters away from the intersection, heading towards B. Vehicle V9110 is in lane 7, moving west at a speed of 20.11 km/h, and is 65.18 meters away from the intersection, heading towards D. | **Conflict Status**: Yes **Conflict Analysis**: A conflict exists between Vehicle V5284 and Vehicle V6460. They approach the intersection from different directions and will reach it at approximately the same moment, creating a potential collision risk. **Recommendations**: To resolve this conflict, Vehicle V6460 should yield to Vehicle V5284, allowing the smaller vehicle to navigate past the intersection first. **Next Actions for Each Vehicle**: - Vehicle V5284: Continue to the intersection as planned - Vehicle V6460: Slow down and prepare to yield to Vehicle V5284 - Vehicle V6758: No action - Vehicle V9110: No action |

## Conclusion

This study presents a new paradigm to traffic control systems, proposing LLMs as dynamic traffic controllers. Leveraging their logical reasoning, scene understanding, and decision-making abilities, LLMs can adaptively respond to real-time traffic conditions and optimize throughput while considering safety, reducing congestion, and environmental impact. Through the 4D traffic control system framework (i.e., Detect, Decide, Disseminate, and Deploy), LLMs centralize processes traditionally managed by disconnected components. LLMs integrate diverse data sources (e.g., loop detectors, GPS, video imaging) to make context-aware decisions and provide tailored control outputs and feedback using diverse modules via various communication channels (e.g., voice, wireless signals, code, and visuals) to drivers, infrastructures, and autonomous vehicles. Their ability to handle multi-objective optimization functions, LLMs can transform urban mobility. We argued in this study that as LLMs offer adaptability in communication, interfacing with humans and machines, this has will potentially ensure compliance and safety in complex traffic scenarios.

We presented a methodology to integrate and evaluate LLMs as a controller into real-time traffic control systems, which comprises of four key stages including data creation and initialization, prompt generation, conflict identification, and fine-tuning with model analysis. Initially, we proposed a system that generated datasets reflecting real-world multi-lane four-leg urban intersection scenarios. We also parsed detailed intersection layouts, lane configurations, and vehicle attributes such as speed, direction, and destination of each vehicle. These datasets are organized into vehicle objects, enabling scenario simulation and conflict detection to find the truth values. Prompts designed using the chain-of-thought method to guide the LLM in analyzing intersection conditions, detect potential conflicts, and suggest priority-based traffic decisions. Conflict resolution contains traditional traffic rules such as the right-hand rule and yielding priorities. These conflict resolutions augmented by LLM's context-aware analyses of vehicle positions, speeds, and arrival times. This multi-stage approach not only enhances conflict detection but also offers adaptive, context-sensitive traffic management solutions that align with real-world regulations. We tested three different LLM models including GPT-mini, Gemini, and Llama. We evaluated the results using accuracy, precision, recall, F1-score, and ROUGE-L.

Results demonstrated that LLM (specifically, a fine-tuned GPT-mini model) has a significant high ability to identify conflicts and support decision-making in traffic intersection scenarios. It is also worth noting that the Gemini and Llama models showed potential for improvement with fine-tuning, however, their performance remained lower than that of the GPT-mini models. This might be likely due to architectural limitations or less effective optimization for this specific task. Nonetheless, the fine-tuned GPT-mini achieved a high accuracy of about 83% and an F1-score of 0.84 in mixed-vehicle scenarios, which reflects its ability to generalize across diverse traffic conditions. This performance is further supported by a high recall and precision rates of about 0.85 and 0.83, respectively. This shows that LLMs can be reliable in detecting actual conflicts, a critical aspect for real-world intersection control applications.

Results also showed the fine-tuned GPT-mini model excelled in generating detailed, actionable insights for decision-making and traffic management. With the help of traffic experts, the LLM model achieved a high ROUGE-L scores across the four key components including conflict identification of about 0.95, decision-making of about 0.91, priority assignment of about 0.94, and waiting times of about 0.92. These results showed that LLM can be practical and applicabile in providing coherent and precise recommendations such as yielding, slowing down, and stopping based on vehicles' speeds, distances, and paths and with considering the other vehicles approaching the intersection.

The findings of this study demonstrated LLMs' capabilities and its potential to transform traffic control systems. It can enhance traffic flow efficiency and safety at intersections. Its ability to handle both simple and complex scenarios makes it a promising tool for real-world traffic management systems in real-time and providing feedback to drivers, infrastructures, and autonomous vehicles. However, future work will focus on further improving performance in high-density traffic scenarios and integrating the model into real-time traffic management applications.

## Author Contributions

Conceptualization, S.M., H.I.A. and M.E.; methodology, S.M., H.I.A., and M.E.; software, S.M.; formal analysis, S.M.; investigation, S.M., and H.I.A.; resources, S.M.; data curation, S.M.; writing—original draft preparation, S.M.; writing—review and editing, H.I.A., and M.E.; visualization, S.M.; supervision, H.I.A., and M.E.; project administration, H.I.A., and M.E. All authors have read and agreed to the published version of the manuscript.

## Funding

This research received no external funding.

## Data Availability Statement

The data and code that support the findings of this study are openly available in Zenodo at https://doi.org/10.5281/zenodo.14171745, reference number 14171745.

## Conflicts of Interest

The authors declare no conflicts of interest.